\documentclass[journal,12pt,onecolumn,draftclsnofoot,]{IEEEtran}
\ifCLASSINFOpdf
\else
\fi
\usepackage{xfrac}
\usepackage{cite}
\usepackage{hyperref}
\usepackage{amsmath}
\usepackage{url}
\usepackage{tabularx}
\usepackage{multirow}
\usepackage{booktabs}
\usepackage{xcolor}
\usepackage{balance}
\usepackage[caption=false]{subfig}
\usepackage{lipsum}
\usepackage{amsfonts}
\usepackage{graphicx}
\usepackage{epstopdf}
\usepackage{algorithm}
\usepackage{algorithmic}
\graphicspath{{figures/}}
\newcommand{\bb}{\boldsymbol}

\begin{document}
\title{Efficient ADMM-based Algorithms for Convolutional Sparse Coding}
\author{Farshad G. Veshki and Sergiy A. Vorobyov
\thanks{The authors are with the Department of Signal Processing and Acoustics, Aalto University, Espoo, Finland (e-mail: farshad.ghorbaniveshki@aalto.fi; sergiy.vorobyov@aalto.fi).}}
\maketitle

\begin{abstract}
Convolutional sparse coding improves on the standard sparse approximation by incorporating a global shift-invariant model. The most efficient convolutional sparse coding methods are based on the alternating direction method of multipliers and the convolution theorem. The only major difference between these methods is how they approach a convolutional least-squares fitting subproblem. This letter presents a solution to this subproblem, which improves the efficiency of the state-of-the-art algorithms. We also use the same approach for developing an efficient convolutional dictionary learning method. Furthermore, we propose a novel algorithm for convolutional sparse coding with a constraint on the approximation error.
\end{abstract}
\IEEEpeerreviewmaketitle

\section{Introduction}
Sparse representations are widely used in various applications of signal and image processing~\cite{face_rec2009,sig_rec2007,im_den20016,im_sr2010,hyper2012,5G2015,my_mffusion,my_mmfusion}. The sparse synthesis model admits that natural signals can be approximated using a linear combination of only a small number of atoms (columns) of a dictionary (matrix). A common formulation of the sparse coding problem is given as
\begin{equation}
\begin{split}
\begin{aligned}
    \underset{\bb{x}}{\mathrm{minimize}} \ \bb{\Gamma}(\bb{x}) \quad {\text{s.t.}} \quad \|\bb{D}\bb{x}-\bb{s}\|_2^2 \leq \epsilon,
\label{eq: standard sparse coding}
\end{aligned} 
\end{split}
\end{equation}
\noindent where $\bb{D}= [ \bb{d}_1,\bb{d}_2, \dots,\bb{d}_K ]$, $\bb{d}_k\in\mathbb{R}^{n}, k=1,\dots,K$, is the dictionary, $\bb{x}\in\mathbb{R}^{m}$ is the vector of sparse coefficients, and $\bb{s}\in\mathbb{R}^{n}$ is the signal. Moreover, $\epsilon$ is the upper bound on the energy of the approximation error and $\bb{\Gamma}(\cdot)$ represents a function that measures the level of sparsity of a vector, for example, \textit{the number of nonzero elements} (denoted by $\|\cdot\|_0$) or its convex relaxation the $\ell_1-$norm (denoted by $\|\cdot\|_1$). The problem of finding sparsity promoting dictionaries is called dictionary learning~\cite{MOD1999,KSVD2006}.  

The applications of sparse representations and dictionary learning usually involve either or both extraction and estimation of local features. Typically, this is handled by a prior decomposition of the original signal into vectorized overlapping blocks (\textit{e.g.}, patches in image processing). As a drawback, this strategy results in multi-valued representations, so that each point in the signal is estimated multiple times. Moreover, since the relationships among neighboring blocks are ignored, dictionaries learned using this approach tend to contain shifted versions of the same features. 

Convolutional sparse coding (CSC) incorporates a single-valued and shift-invariant model that represents the entire signal. In this model, the product $\bb{D}\bb{x}$ in the standard sparse coding problem is replaced by a sum of convolutions. The convolutional form of the standard sparse coding problem \eqref{eq: standard sparse coding} can be written as follows
\begin{equation}
\begin{split}
\begin{aligned}
    \underset{\{\bb{x}_k\}_{k=1}^K}{\mathrm{minimize}} \ \sum_{k=1}^K\bb{\Gamma}(\bb{x}_k) \quad {\text{s.t.}} \quad \Big\|\sum_{k=1}^K\bb{d}_k\ast\bb{x}_k-\bb{s}\Big\|_2^2 \leq \epsilon,
\label{eq: CSC general}
\end{aligned} 
\end{split}
\end{equation}
where $\ast$ denotes the convolution operator (usually, with a ``$\textit{same}$'' padding), and $\boldsymbol{x}_k \in \mathbb{R}^{n}$ and $\boldsymbol{d}_k \in \mathbb{R}^{m}, \ k=1,\cdots,K$, are the sparse coefficient maps and the dictionary filters, respectively. Several applications have shown that the CSC model performs better in handling natural signals, such as audio and images, in comparison with its standard version~\cite{piano2017,music2016,CT2019,fusion2016,RGB2018,superres2015,rain2018,astro2021}.

A majority of available CSC algorithms, including~\cite{Bristow2013,Heide2015,Wohlberg2016,Peng2019,Choudhury2017,Papyan2017,Otero2020,Papyan20172,Wang2018}, are based on the alternating direction method of multipliers (ADMM) framework~\cite{ADMM2011}. ADMM breaks the CSC problem into two main sub-problems, one of which is a sparse approximation problem which is efficiently addressed using hard-thresholding (when $\bb{\Gamma}(\bb{x}) = \|\bb{x}\|_0$) or a shrinkage operator (when $\bb{\Gamma}(\bb{x}) = \|\bb{x}\|_1$), and the other entails a convolutional least-squares regression. An efficient solution to the second sub-problem based on the convolution theorem and the Sherman-Morrison formula is given in \cite{Wohlberg2016}. CSC problem \eqref{eq: CSC general} is typically addressed by solving its unconstrained equivalent, which is written as
\begin{equation}
\begin{split}
\begin{aligned}
    \underset{\{\bb{x}_k\}_{k=1}^K}{\mathrm{minimize}} \ \frac{1}{2}\Big\|\sum_{k=1}^K\bb{d}_k\ast\bb{x}_k-\bb{s}\Big\|_2^2 + \lambda \sum_{k=1}^K\bb{\Gamma}(\bb{x}_k) ,
\label{eq: unconstrained CSC general}
\end{aligned} 
\end{split}
\end{equation}
\noindent where $\lambda>0$ is a Lagrange multiplier. It is known that there is a unique $\lambda$ for each $\epsilon$. However, the appropriate value of $\lambda$ also depends on $\boldsymbol{s}$ and $\{\bb{d}_k\}_{k=1}^K$. Thus, despite being more convenient to solve, the unconstrained reformulation introduces data dependency to the CSC algorithm. 

A common approach for convolutional dictionary learning (CDL) entails optimizing the filters and the sparse coefficient maps using a batch of $P$ training signals~\cite{Heide2015,Choudhury2017,Wohlberg2016,Peng2019}. This problem can be formulated as follows
\begin{equation}
\begin{split}{
\begin{aligned}
    &\underset{\{\bb{x}_k^p\}_{k=1}^K,\{\bb{d}_k\}_{k=1}^K}{\mathrm{minimize}} \ \sum_{p=1}^P\Bigg(\frac{1}{2}\Big\|\sum_{k=1}^K\bb{d}_k\ast\bb{x}_k^p-\bb{s}^p\Big\|_2^2 + \lambda \sum_{k=1}^K\bb{\Gamma}(\bb{x}_k^p)\Bigg)\\ 
    &\quad\text{s.t.} \quad \bb{d}_k \in \bb{\mathcal{D}}
\label{eq: CDL genaral}
\end{aligned} }
\end{split}
\end{equation}
where $\bb{\mathcal{D}} = \{ \boldsymbol{d}_k \in \mathbb{R}^{m} \ | \ \|\bb{d}\|_2 =1, k =1, \dots, K\}$. The CDL problem is usually addressed by alternating optimization with respect to $\{\bb{x}_k^p\}_{k=1}^K$ and $\{\bb{d}_k\}_{k=1}^K$\cite{Bristow2013,Heide2015,Wohlberg2016}. Several works have shown that solving \eqref{eq: CDL genaral} with respect to $\{\bb{d}_k\}_{k=1}^K$ can be also done effectively and efficiently using ADMM in frequency domain~\cite{cardona2018}.

This paper presents a direct method for solving the convolutional least-squares regression which yields a constant improvement on the complexity of the available CSC algorithms. The same method can be used to improve the efficiency of existing CDL methods. Additionally, we propose an efficient CSC algorithm with a constraint on the energy of the approximation residuals using our solution to the unconstrained CSC problem. MATLAB implementations of the proposed algorithms are available at GitHub repository \cite{GithubRep}.

Throughout the paper, we use $(\cdot)^T$ to denote the (non-conjugate) transpose operator. $(\bar{\cdot})$ represents complex-conjugate of complex number, $(\hat{\cdot})$ denotes the discrete Fourier transform of a signal, and $({\cdot})^\star$ denotes the solution to an optimization problem. Moreover, we use $\odot$ and $\oslash$ to denote element-wise multiplication and element-wise division operators, respectively.
\section{Proposed Algorithms}
\subsection{Unconstrained CSC}
\label{sec: unconstrained CSC}
In this work, we consider the convex formulation of CSC problem, \textit{i.e.}, we use $\bb{\Gamma}(\bb{x}) = \|\bb{x}\|_1$. Using variable splitting, problem~\eqref{eq: unconstrained CSC general} in ADMM form can be reformulated as~\cite{ADMM2011}, 
\begin{equation}
\begin{split}
\begin{aligned}
    &\underset{\{\bb{x}_k\}_{k=1}^K}{\mathrm{minimize}} \ \frac{1}{2}\Big\|\sum_{k=1}^K\bb{d}_k\ast\bb{z}_k-\bb{s}\Big\|_2^2 + \lambda \sum_{k=1}^K \|\bb{x}_k\|_1\\
    &\text{s.t.} \quad \bb{z}_k =\bb{x}_k, \ k=1,\dots,K.
\label{eq: ADMM unconstrained CSC}
\end{aligned} 
\end{split}
\end{equation}
The augmented Lagrangian corresponding to \eqref{eq: ADMM unconstrained CSC} is written as
{\begin{equation}
\frac{1}{2}\Big\|\sum_{k=1}^K\bb{d}_k\ast\bb{z}_k-\bb{s}\Big\|_2^2 + \lambda \sum_{k=1}^K \|\bb{x}_k\|_1+ \sum_{k=1}^K \bb{y}_k^T(\bb{z}_k-\bb{x}_k) + \frac{\rho}{2}\sum_{k=1}^K\|\bb{z}_k-\bb{x}_k\|_2^2.
\label{eq: unconstrained augmented laugrangian}
\end{equation}}
\noindent where $\rho>0$ is the penalty parameter and $\{\bb{y}_k\}_{k=1}^K$ are Lagrangian multipliers. Defining $\bb{u}_k = (\sfrac{1}{\rho})\bb{y}_k$, the scaled-form ADMM iterations are expressed as
\begin{equation}{
\begin{aligned}
    &\{\bb{z}_k^{t+1}\}_{k=1}^K = \underset{\{\bb{z}_k\}_{k=1}^K} {\mathrm{argmin}} \left( \frac{1}{2}\Big\|\sum_{k=1}^K\bb{d}_k \ast\bb{z}_k \!-\! \bb{s} \Big\|_2^2 + \frac{\rho}{2}\sum_{k=1}^K\|\bb{z}_k - \bb{x}_k^t + \bb{u}_k^t\|_2^2 \right)\\
    &\{\bb{x}_k^{t+1}\}_{k=1}^K = \underset{\{\bb{x}_k\}_{k=1}^K} {\mathrm{argmin}} \left( \lambda \sum_{k=1}^K \|\bb{x}_k\|_1 +  \frac{\rho}{2}\sum_{k=1}^K\|\bb{z}_k^{t+1}-\bb{x}_k + \bb{u}_k^t\|_2^2 \right)\\
    &\bb{u}_k^{t+1} = \bb{u}_k^{t} + \bb{z}_k^{t+1} - \bb{x}_k^{t+1}, \qquad k = 1,\dots,K.
\label{eq: ADMM iterations unconstrained CSC L1}
\end{aligned}}
\end{equation}
The second subproblem ($\bb{x}$-update step) can be addressed in an element-wise manner using a shrinkage (soft-thresholding) operator. The solution is written as
\begin{equation}
    {{\boldsymbol{x}_k}}^{t+1} = \mathcal{S}_{\sfrac{\lambda}{\rho}}\big({{\boldsymbol{z}_k}}^{t+1} + {{\boldsymbol{u}_k}}^{t}\big) \label{eq: subprolem 2 shrinkage}
\end{equation}
with the shrinkage operator defined as follows
\begin{equation}
    \mathcal{S}_{\kappa}(a) = \mathrm{sign}(a) \mathrm{max}(0,|a|-\kappa).\label{eq: shrinkage}
\end{equation}

The only challenging step is solving the first subproblem ($\bb{z}$-update step). In a general form, this step entails solving the optimization problem 
\begin{equation}
   \underset{ \{\boldsymbol{z}_k\}_{k=1}^K  }{\mathrm{minimize}} \ \frac{1}{2}\|\sum_{k=1}^K \boldsymbol{d}_k \ast \boldsymbol{z}_k -  \boldsymbol{s}\|_2^2 + \frac{\rho}{2} \sum_{k=1}^K\|\boldsymbol{z}_k-\boldsymbol{w}_k\|_2^2. \label{eq: subproblem X}
\end{equation} 
Using the convolution theorem, problem \eqref{eq: subproblem X} in Fourier domain can be written as
\begin{equation}
   \underset{ \{\hat{\boldsymbol{z}}_k\}_{k=1}^K  }{\mathrm{minimize}} \ \frac{1}{2}\|\sum_{k=1}^K \hat{\boldsymbol{d}}_k \odot \hat{\boldsymbol{z}}_k -  \hat{\boldsymbol{s}}\|_F^2 + \frac{\rho}{2} \sum_{k=1}^K \|\hat{\boldsymbol{z}}_k-\hat{\boldsymbol{w}}_k\|_2^2. \label{eq: subproblem X fourier}
\end{equation}
Note that the filters $\{\bb{d}_k\}_{k=1}^K$ are zero-padded to the size of $\{\bb{z}_k\}_{k=1}^K$ before performing the discrete Fourier transform. Denoting 
\begin{equation}
\begin{aligned}
    &\boldsymbol{\delta}_i = [ \hat{\boldsymbol{d}}_1(i), \cdots, \hat{\boldsymbol{d}}_k(i) ]^T,\\
    &\boldsymbol{\zeta}_i = [ \hat{\boldsymbol{z}}_1(i), \cdots, \hat{\boldsymbol{z}}_k(i) ]^T,\\
    &\boldsymbol{\omega}_i = [ \hat{\boldsymbol{w}}_1(i), \cdots, \hat{\boldsymbol{w}}_k(i) ]^T, \label{eq: alpha beta omega}
\end{aligned}
\end{equation}
with $i = 1,\dots,n$, problem~\eqref{eq: subproblem X fourier} can be addressed as $n$ independent problems:
\begin{equation}
   \underset{ {\boldsymbol{\zeta}}_i  }{\mathrm{minimize}} \ \label{eq: subproblem X vectorized} \frac{1}{2}\big({{\boldsymbol{\delta}}}_{i}^T {\boldsymbol{\zeta}}_{i} -  \hat{{s}}_{i}\big)^2 + \frac{\rho}{2} \|{\boldsymbol{\zeta}}_{i}-{\boldsymbol{\omega}}_{i}\|_2^2. 
\end{equation}
Equating the derivative with respect to ${\boldsymbol{\zeta}}_i$ to zero, we have
\begin{equation}{
\begin{aligned}
    0 & = \bar{\boldsymbol{\delta}}_{i}({{\boldsymbol{\delta}}}_{i}^T {\boldsymbol{\zeta}}_{i} -  \hat{{s}}_{i}\big) + \rho{\boldsymbol{\zeta}}_{i} - \rho{\boldsymbol{\omega}}_{i} \\
     & =  (\bar{\boldsymbol{\delta}}_{i}{{\boldsymbol{\delta}}}_{i}^T+\rho\boldsymbol{I}){\boldsymbol{\zeta}}_{i} - \hat{{s}}_{i}\bar{\boldsymbol{\delta}}_{i} - \rho{\boldsymbol{\omega}}_{i}\\
     & =  (\bar{\boldsymbol{\delta}}_{i}{{\boldsymbol{\delta}}}_{i}^T+\rho\boldsymbol{I}){\boldsymbol{\zeta}}_{i} - (\hat{{s}}_{i}\bar{\boldsymbol{\delta}}_{i} - \bar{\boldsymbol{\delta}}_{i}{{\boldsymbol{\delta}}}_{i}^T{\boldsymbol{\omega}}_{i})- (\bar{\boldsymbol{\delta}}_{i}{{\boldsymbol{\delta}}}_{i}^T+\rho\boldsymbol{I}){\boldsymbol{\omega}}_{i} \label{eq: z-update derivation}
\end{aligned}} 
\end{equation}
which gives
\begin{equation}
    \begin{split}
        \begin{aligned}
    {\boldsymbol{\zeta}}_i^{\star} & = \boldsymbol{\omega}_i + {(\hat{\boldsymbol{s}}_i -{\boldsymbol{\delta}}_i^T\boldsymbol{\omega}_i)}{(\bar{\boldsymbol{\delta}}_{i}{{\boldsymbol{\delta}}}_{i}^T+\rho\boldsymbol{I})^{-1}}\bar{\boldsymbol{\delta}}_i\\
    & =  \boldsymbol{\omega}_i + {(\hat{\boldsymbol{s}}_i -{\boldsymbol{\delta}}_i^T\boldsymbol{\omega}_i)}{(\|\boldsymbol{\delta}_i\|_2^2+\rho )^{-1}}\bar{\boldsymbol{\delta}}_i.
        \end{aligned}
   \end{split}\label{eq: optimal x}
\end{equation} 

Denoting
\begin{equation}{
\begin{aligned}
 \hat{\boldsymbol{c}}_k^{\rho}={\bar{\hat{\boldsymbol{d}}}_k}\oslash
    \big({\rho+\sum_{k=1}^K{\bar{\hat{\boldsymbol{d}}}_k\odot\hat{\boldsymbol{d}}_k}\big) } \quad \text{and} \quad  \hat{\boldsymbol{r}}=\hat{\boldsymbol{s}}-\sum_{k=1}^K\hat{\boldsymbol{d}}_k\odot\hat{\boldsymbol{w}}_k,
    \end{aligned}}
    \label{eq: r and c}
\end{equation}
\noindent the solution to the $\bb{z}$-update step based on \eqref{eq: optimal x} can be written as 
\begin{equation}
    \begin{split}
        \begin{aligned}
        {\hat{\boldsymbol{z}}_k^{\star} = \hat{\boldsymbol{w}}_k + \hat{\boldsymbol{c}}_k^{\rho} \odot \hat{\boldsymbol{r}}}.
        \end{aligned}
   \end{split}\label{eq: X update}
\end{equation}
\subsubsection*{Computational Complexity}
The available ADMM-based CSC algorithms usually address the $\bb{z}$-update step by computing the following
\begin{equation}{
\begin{aligned}
    {\boldsymbol{\zeta}}_{i}^{\star} = (\bar{\boldsymbol{\delta}}_{i}{{\boldsymbol{\delta}}}_{i}^T+\rho\boldsymbol{I})^{-1}(\hat{{s}}_{i}\bar{\boldsymbol{\delta}}_{i} + \rho{\boldsymbol{\omega}}_{i}), \label{eq: x_update inv}
\end{aligned}} 
\end{equation}
which can be inferred from the second line of \eqref{eq: z-update derivation}. Solving problem \eqref{eq: x_update inv} using direct matrix inversion results in a time complexity of $\bb{\mathcal{O}}(K^3)$ \cite{Bristow2013}. However, the work of \cite{Wohlberg2016} demonstrated that this can be reduced to $\bb{\mathcal{O}}(K)$ using the Sherman-Morrison formula. The time complexity of the proposed method is also of $\bb{\mathcal{O}}(K)$. However, using further simplifications, the proposed approach eliminates the need for explicit matrix inversion and requires fewer computations. In particular, performing the $\bb{z}$-update step on a batch of $P$ images using the proposed method requires $((4K+1)P+3K+1)n$ flops, while it takes $(7KP+3K+1)n$ flops using the method of \cite{Wohlberg2016}, indicating a considerable improvement provided by our method.

\subsection{Constrained CSC}
The ADMM formulation of the constrained CSC problem \eqref{eq: CSC general} is given as
\begin{equation}
\begin{split}
\begin{aligned}
    &\underset{\{\bb{x}_k\}_{k=1}^K}{\mathrm{minimize}} \ \bb{f}\big(\{\bb{z}_k\}_{k=1}^K \big) + \sum_{k=1}^K \|\bb{x}_k\|_1\\
    &\text{s.t.} \quad \bb{z}_k =\bb{x}_k, \ k=1,\dots,K
\label{eq: constrained CSC ADMM}
\end{aligned} 
\end{split}
\end{equation}
where $\bb{f}\big(\{\bb{z}_k\}_{k=1}^K \big)$ is an indicator function of the constraint set in \eqref{eq: unconstrained CSC general}, that is,
\begin{equation}{\small
    \bb{f}\big(\{\bb{z}_k\}_{k=1}^K \big) = 
     \begin{cases}
       0, &\quad\text{\textit{if}}\quad \bb{e}\big(\{\bb{z}_k\}_{k=1}^K \big) \leq \epsilon\\
       \infty, &\quad\text{\textit{o.w.}} \\ 
     \end{cases}},
\end{equation}
where
\begin{equation}
    \bb{e}\big(\{\bb{z}_k\}_{k=1}^K \big) = \Big\|\sum_{k=1}^K\bb{d}_k\ast\bb{z}_k-\bb{s}\Big\|_2^2.\label{eq: error}
\end{equation}
The ADMM iterations are
\begin{equation}{
\begin{aligned}
    &\{\bb{z}_k^{t+1} \}_{k=1}^K = \underset{\{\bb{z}_k\}_{k=1}^K}
    {\mathrm{argmin}} \left(\bb{f}\left( \{\bb{z}_k\}_{k=1}^K \right) +  \frac{\rho}{2}\sum_{k=1}^K \| \bb{z}_k - \bb{x}_k^t + \bb{u}_k^t\|_2^2 \right)\\
    &\{\bb{x}_k^{t+1}\}_{k=1}^K  = \underset{\{\bb{x}_k\}_{k=1}^K} {\mathrm{argmin}} \left( \sum_{k=1}^K \|\bb{x}_k\|_1 +  \frac{\rho}{2}\sum_{k=1}^K \|\bb{z}_k^{t+1}-\bb{x}_k + \bb{u}_k^t\|_2^2 \right)\\
    &\bb{u}_k^{t+1} = \bb{u}_k^{t} + \bb{z}_k^{t+1} - \bb{x}_k^{t+1}, \qquad k = 1,\dots,K.
\label{eq: ADMM iterations constrained}
\end{aligned}}
\end{equation}
The $\bb{z}$-update step requires solving the following optimization problem
\begin{equation}
   \underset{ \{\boldsymbol{z}_k\}_{k=1}^K  }{\mathrm{minimize}} \  \bb{f}\big(\{\bb{z}_k\}_{k=1}^K \big) + \frac{\rho}{2} \sum_{k=1}^K\|\boldsymbol{z}_k-\boldsymbol{w}_k\|_2^2. \label{eq: subproblem X constrained}
\end{equation} 
Depending on $\{\boldsymbol{w}_k\}_{k=1}^K$, problem \eqref{eq: subproblem X constrained} either has a trivial solution or it is equivalent to an equality-constrained optimization problem. This can be expressed as
\begin{equation}
    \begin{aligned}{
    \{\bb{z}_k^{\star}\}_{k=1}^K=
     \begin{cases}
       \{\bb{w}_k\}_{k=1}^K, \quad &\text{\textit{if}}\quad \bb{e}\big(\{\bb{w}_k\}_{k=1}^K \big) \leq \epsilon\\
        \scriptsize{\underset{ \{\boldsymbol{z}_k\}_{k=1}^K  }{\mathrm{argmin}}\sum\limits_{k=1}^{K}\|\boldsymbol{z}_k-\boldsymbol{w}_k\|_2^2 \quad\text{s.t.} \ \bb{e}\big(\{\bb{z}_k\}_{k=1}^K\big)=\epsilon},  &\text{\rm{otherwise}} \\ 
     \end{cases}}.
     \end{aligned}\label{eq: constrained CSC conditions}
\end{equation}
Using a suitable Lagrange multiplier $\nu$, the problem in the second term of \eqref{eq: constrained CSC conditions} can be reformulated as
\begin{equation}
   \underset{ \{\boldsymbol{z}_k\}_{k=1}^K  }{\mathrm{minimize}} \ \bb{e}\big(\{\bb{z}_k\}_{k=1}^K\big) + {\nu} \sum_{k=1}^K\|\boldsymbol{z}_k-\boldsymbol{w}_k\|_2^2, \label{eq: constrained CSC equality term}
\end{equation}
which has the same form as problem \eqref{eq: subproblem X}. Finding the solution of \eqref{eq: constrained CSC equality term} using \eqref{eq: X update} and plugging it into \eqref{eq: error} gives
\begin{equation}
\begin{aligned}{
    \bb{e}\big(\{\bb{z}_k^{\star}\}_{k=1}^K\big)= \frac{\nu^2}{n} \Bigg\|\hat{\boldsymbol{r}}\oslash\big(\nu+\sum_{k=1}^K{\bar{\hat{\boldsymbol{d}}}_k\odot\hat{\boldsymbol{d}}_k}\big)\Bigg\|_2^2,
}\end{aligned}
\end{equation}
where the division by $n$ is required by Parseval's theorem. Thus, problem \eqref{eq: subproblem X constrained} is simplified to a single-variable optimization problem for finding the optimal multiplier $\nu^{\star}$, which satisfies
\begin{equation}
    \nu^{\star} = \Big\{\nu \ | \  \bb{e}\big(\{\bb{z}_k^{\star}\}_{k=1}^K\big) = \epsilon \Big\}.
\end{equation}
Considering that $\bb{e}\big(\{\bb{z}_k^{\star}\}_{k=1}^K\big)$ is monotonically increasing in $\nu~>~0$, this problem can be efficiently addressed, for example, using the \textit{secant} method. Once $\nu^{\star}$ is known, the $\bb{z}$-update can be performed as  
\begin{equation}
    \begin{split}
        \begin{aligned}
        {\hat{\boldsymbol{z}}_k^{\star} = \hat{\boldsymbol{w}}_k + \hat{\boldsymbol{c}}_k^{\nu^{\star}} \odot \hat{\boldsymbol{r}}}, \qquad k=1, \cdots, K
        \end{aligned}
   \end{split}\label{eq: X update constrained}
\end{equation}
where $\hat{\boldsymbol{c}}_k^{\nu^{\star}}$ and $\hat{\boldsymbol{r}}$ are calculated using \eqref{eq: r and c}.

\subsection{Dictionary Update}
\label{sec: dictionary update}
Addressing CDL optimization problem \eqref{eq: CDL genaral} over $\{\bb{d}_k\}_{k=1}^K$ is equivalent to solving the following optimization problem 
\begin{equation}
\begin{split}{
\begin{aligned}
    \underset{\{\bb{d}_k\}_{k=1}^K}{\mathrm{minimize}} \ \frac{1}{2}\sum_{p=1}^P\Big\|\sum_{k=1}^K\bb{d}_k\ast\bb{x}_k^p-\bb{s}^p\Big\|_2^2 + \sum_{k=1}^K\boldsymbol{\Omega}(\bb{d}_k) 
\label{eq: CDL step}
\end{aligned} }
\end{split}
\end{equation}
\noindent where $\boldsymbol{\Omega}(\bb{d}_k)$ is an indicator function associated with the constraint set in \eqref{eq: CDL genaral}. Problem \eqref{eq: CDL step} can be efficiently addressed using the consensus ADMM method~\cite{cardona2018}. The consensus ADMM formulation of problem \eqref{eq: CDL step} is given as
\begin{equation}
\begin{split}{
\begin{aligned}
    &\underset{\{\bb{d}_k\}_{k=1}^K}{\mathrm{minimize}} \ \frac{1}{2}\sum_{p=1}^P\Big\|\sum_{k=1}^K\bb{g}_k^p\ast\bb{x}_k^p-\bb{s}^p\Big\|_2^2 + \sum_{k=1}^K\boldsymbol{\Omega}(\bb{d}_k)\\ 
    &\quad\text{s.t.} \quad \bb{g}_k^p =\bb{d}_k, \quad \forall k,p.
\label{eq: CDL step consensus}
\end{aligned} }
\end{split}
\end{equation}
with the ADMM iterations
\begin{equation}{
\begin{aligned}
    &\{ \bb{g}_k^{p,{t+1}} \}_{k=1}^K  = \underset{\{\bb{g}_k^p\}_{k=1}^K }{\mathrm{argmin}} \left( \frac{1}{2}\left\| \sum_{k=1}^K\!\bb{g}_k^p\!\ast\!\bb{x}_k^p - \bb{s}^p\right\|_2^2 + \frac{\sigma}{2}\sum_{k=1}^K\|\bb{g}_k^p - \bb{d}_k^t + \bb{v}_k^{p,t}\|_2^2 \right)\\
    &\{ \bb{d}_k^{t+1} \}_{k=1}^K = \underset{\{\bb{d}_k\}_{k=1}^K } {\mathrm{argmin}} \left( \sum_{k=1}^K\boldsymbol{\Omega}(\bb{d}_k) +  \frac{\sigma}{2}\!\sum_{k=1}^K\|\bb{d}_k - \frac{1}{P}\sum_{p=1}^P(\bb{g}_k^{p,t+1} + \bb{v}_k^{p,t})\|_2^2 \right)\\
    &\bb{v}_k^{p,t+1} = \bb{v}_k^{p,t} + \bb{g}_k^{p,t+1} - \bb{d}_k^{t+1}, \quad k = 1,\dots,K, \quad p = 1,\dots,P.
\label{eq: CDL ADMM iterations}
\end{aligned}}
\end{equation}

The first subproblem ($\bb{g}$-update) is similar to problem \eqref{eq: subproblem X}. Thus, it can be efficiently  addressed using the proposed approach in Section~\ref{sec: unconstrained CSC}. The use of the Fourier domain-based approach requires  $\{\bb{g}_k^p\}_{k=1}^K$ to be the same size as $\{\bb{x}_k^p\}_{k=1}^K$. As a result, the filters $\{\bb{d}_k\}_{k=1}^K$ are zero-padded to the size of $\{\bb{x}_k^p\}_{k=1}^K$ to be conformable with $\{\bb{g}_k^p\}_{k=1}^K$. The second subproblem ($\bb{d}_k$-update) can be solved simply by projecting $\frac{1}{P}\sum_{p=1}^P(\bb{g}_k^{p,t+1} + \bb{v}_k^{p,t})$ on the constraint set by mapping the entries outside the constraint support to zero before normalizing the $\ell_2-$norm.

\subsection{CDL Algorithm}
\label{sec: CDL algorithm}
CDL problem \eqref{eq: CDL genaral} is addressed by alternating between CSC (see Section~\ref{sec: unconstrained CSC}) and dictionary update (see Section~\ref{sec: dictionary update}) subproblems. We use a single iteration for each subproblem. This approach has been shown to be effective while simplifying the algorithm~\cite{Wohlberg2016,cardona2018}. We also use the variable coupling approach suggested in \cite{cardona2017} which is shown to provide a better numerical stability~\cite{Wohlberg2016,cardona2018}. Specifically, the sparse codes $\{\bb{x}_k^p\}_{k=1}^K$ and the constrained filters $\{\bb{d}_k\}_{k=1}^K$ are passed to the next subproblem. 
\section{Experimental Results}
In this section, we first compare the proposed unconstrained CSC algorithm with the state-of-the-art method, which uses the Sherman-Morrison formula in convolutional fitting step (the SM method)~\cite{Wohlberg2016}. Then, we compare our unconstrained and constrained CSC methods in terms of convergence speed. Finally, we compare the proposed CDL algorithm with three available methods. All methods are based on the same alternating approach explained in Section~\ref{sec: CDL algorithm} and use ADMM in both phases (CSC and dictionary update). All compared methods use the SM method in CSC phase. The compared dictionary learning methods are based on the conjugate gradient method (CG)~\cite{Wohlberg2016}, the iterative Sherman-Morrison method (ISM)~\cite{Wohlberg2016} and a method based on the consensus ADMM framework and the Sherman-Morrison formula (SM-cns)~\cite{cardona2018}. 

A $512\times512$ greyscale Lena image is used in the CSC experiments. The CDL experiments are performed using a dataset of 20 images taken from the USC-SIPI database~\cite{SIPIdatabase}. All images in the dataset are converted to greyscale and resized to $256\times256$ pixels. All methods are implemented using MATLAB. All experiments are conducted on a PC equipped with an Intel(R) Core(TM) i5-8365U 1.60GHz CPU.
\subsection{CSC Results}
\begin{figure}[tbhp]
\centering
\subfloat[]{\includegraphics[width=0.5\linewidth]{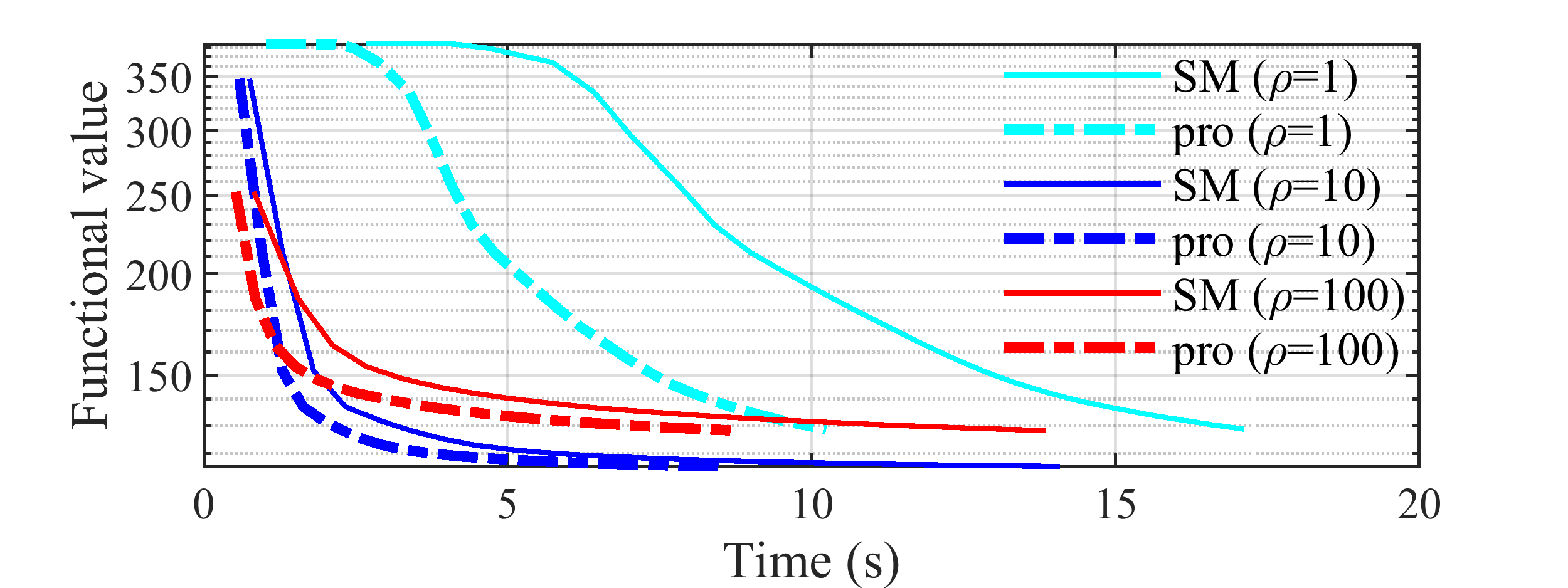}}\\
\subfloat[]{\includegraphics[width=0.5\linewidth]{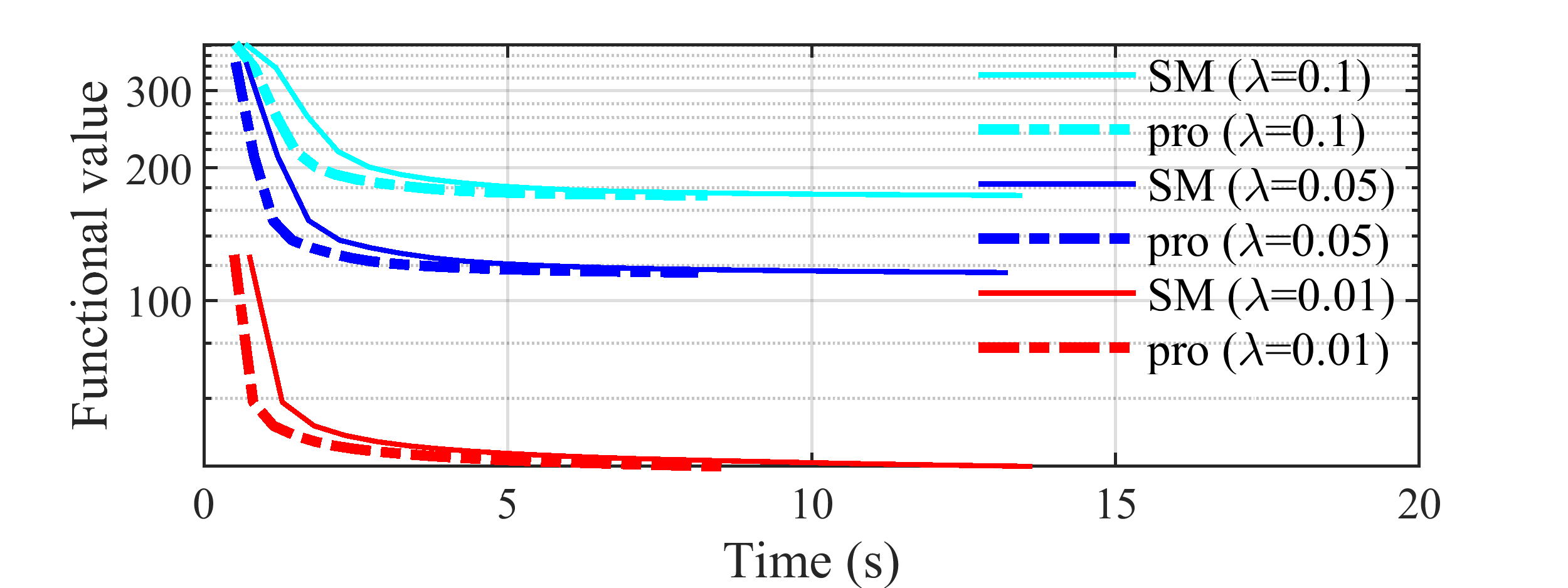}}\\
\caption{Functional values over time for the proposed unconstrained CSC method and the SM method using (a) different values of $\rho$ for $\lambda=0.05$, and (b) different values of $\lambda$ for $\rho=10$. A dictionary of $16$ filters of size $8\times8$ is used in both cases.}
\label{fig: CSC results}
\end{figure}

Fig.~\ref{fig: CSC results} shows the functional values over time for 25 iterations of the proposed unconstrained CSC method and the SM method using different values of $\rho$ and $\lambda$. We use a fixed number of iterations to display the deference in efficiencies (the iterations of the two methods are equally effective). As it can be seen, the proposed method is significantly more efficient in all cases. The algorithm complexities have been compared in Section~\ref{sec: unconstrained CSC}.

\begin{figure}[tbhp]
\centering
{\includegraphics[width=0.5\linewidth]{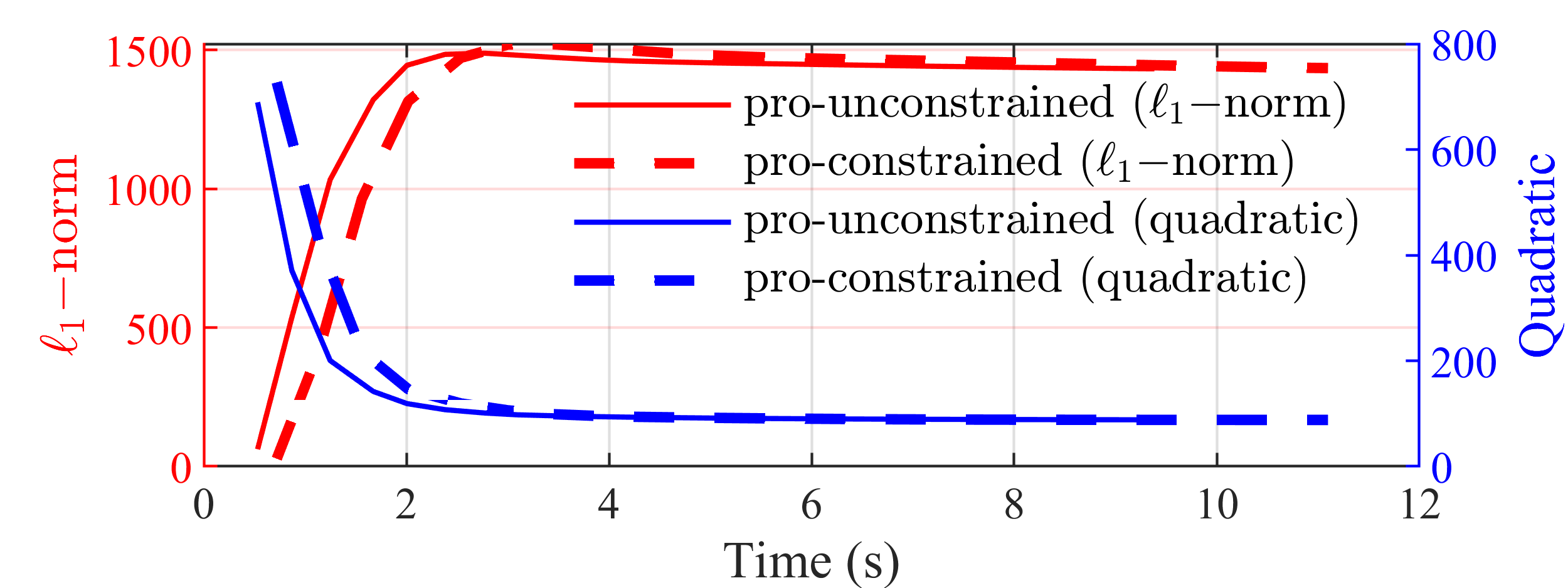}}\\
\caption{The quadratic and  $\ell_1-$norm functional values for the proposed unconstrained and constrained CSC methods using $\lambda=0.05$ ($\epsilon=88.1886$), $\rho=10$. A dictionary of $16$ filters of size $8\times8$ is used.}
\label{fig: CSC results constrained}
\end{figure}

The proposed constrained and unconstrained CSC methods are compared in Fig.~\ref{fig: CSC results constrained}. Specifically, we executed the unconstrained CSC method using $\lambda=0.05$, then we used the observed quadratic functional value ($\epsilon=88.1886$) to run our constrained CSC method, while keeping the rest of the parameters unchanged. As it can be seen, the quadratic and the $\ell_1-$norm functionals converge to the same values for both CSC methods. The constrained method results in a longer runtime, which accounts for optimization with respect to $\nu$ in each iteration.
\subsection{CDL Results}
\begin{figure}[tbhp]
\centering
\subfloat[$P=1$]{\includegraphics[width=0.5\linewidth]{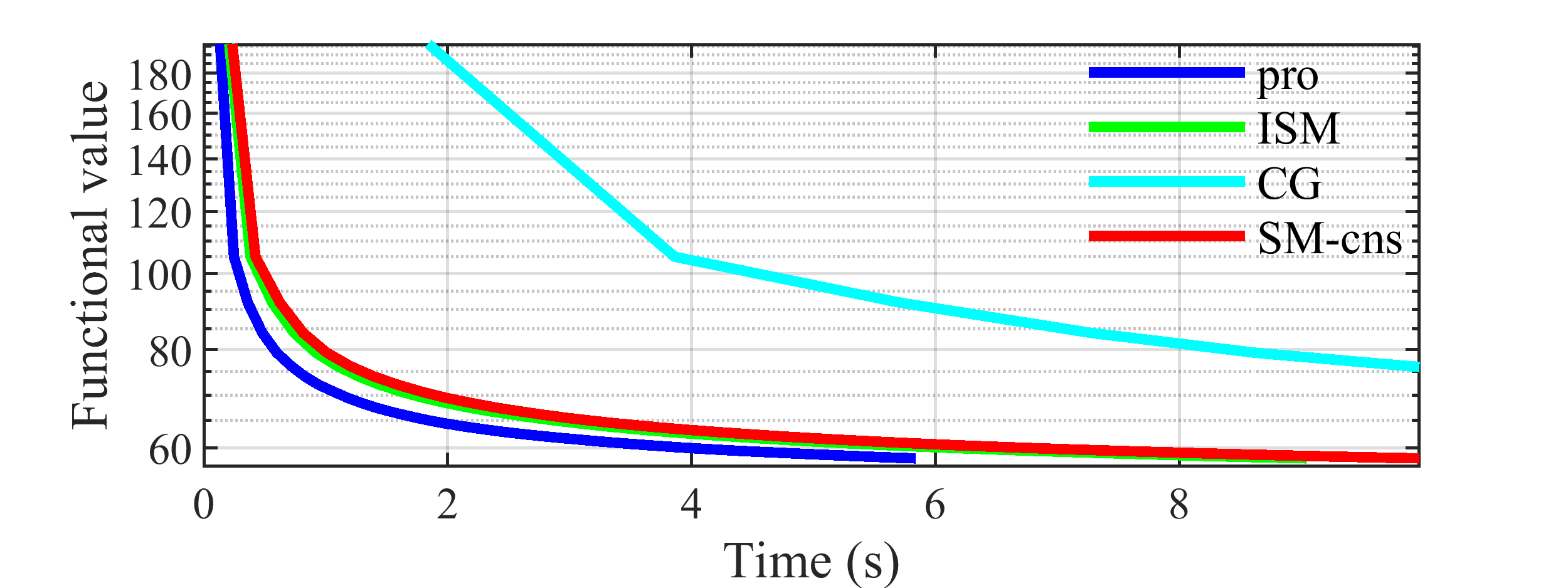}}\\
\subfloat[$P=10$]{\includegraphics[width=0.5\linewidth]{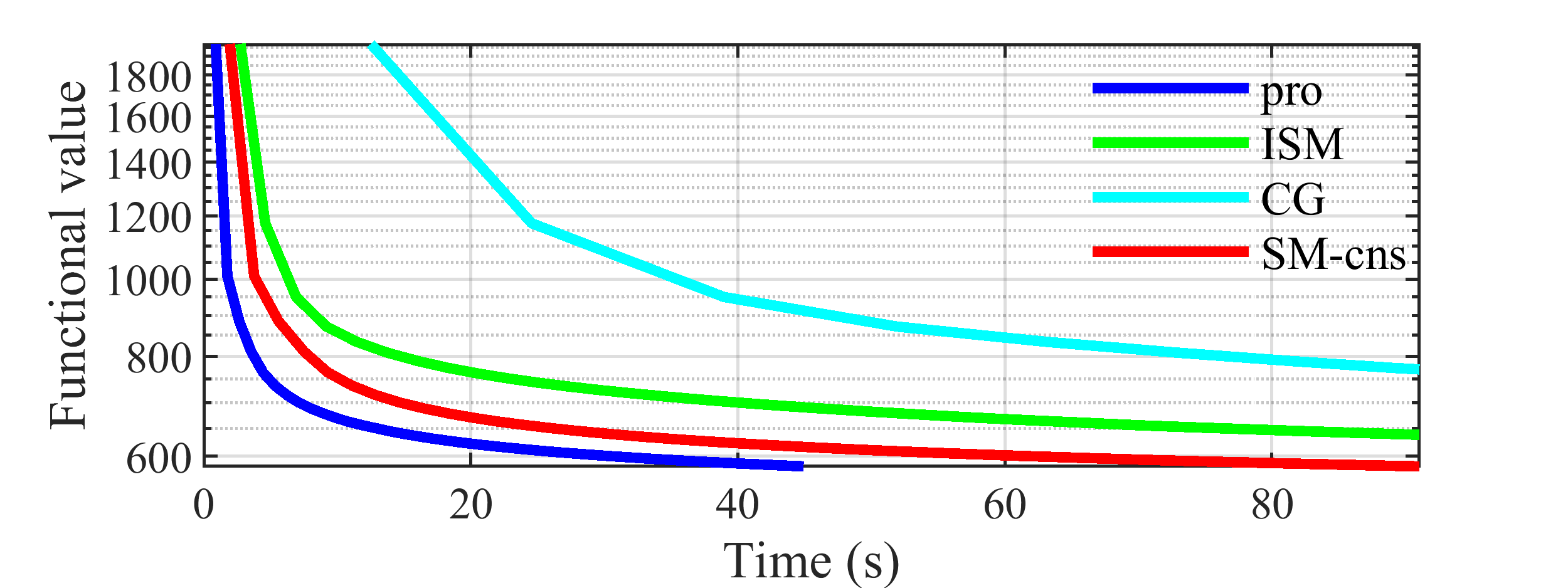}}\\
\subfloat[$P=20$]{\includegraphics[width=0.5\linewidth]{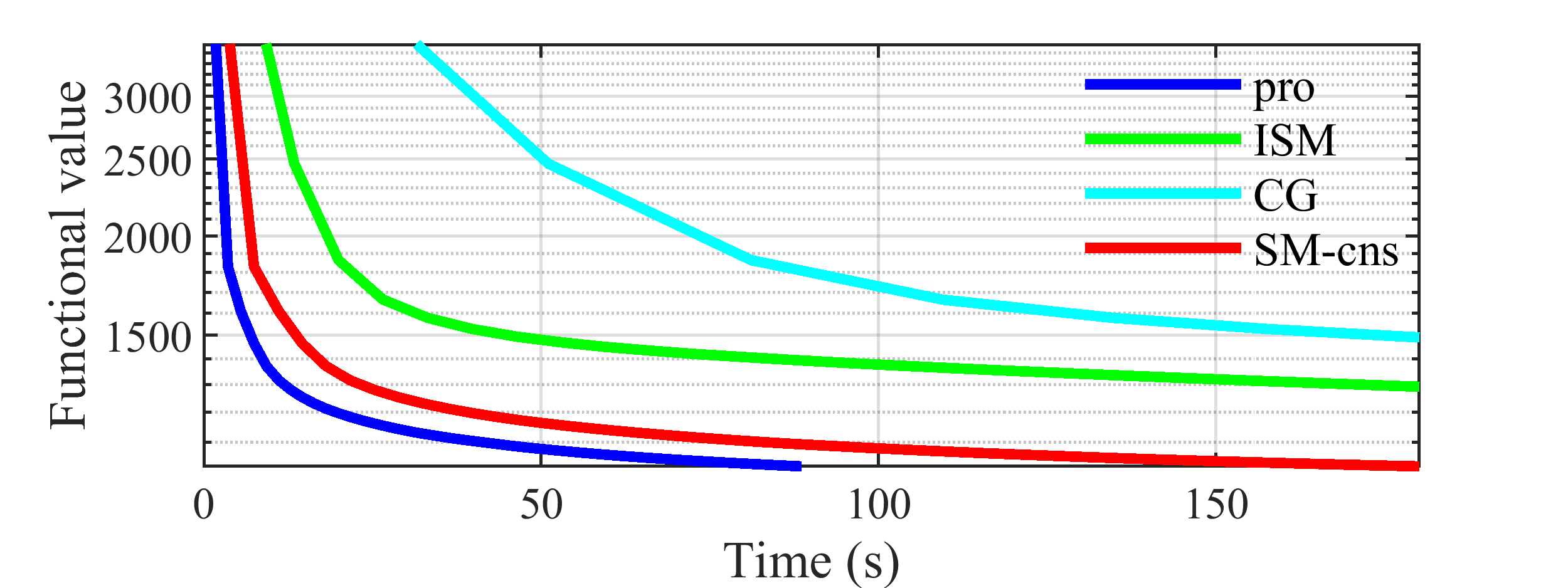}}\\
\caption{Functional values over time using different values of $P$, $\rho=10$, $\lambda=0.05$, $K=16$ filters of size $8\times8$ .}
\label{fig: CDL results}
\end{figure}

In Fig.~\ref{fig: CDL results}, the functional values over time for 50 iterations of all CDL methods using different dataset sizes ($P$) are compared. The complexity of the ISM method is of $\bb{\mathcal{O}}(KP^2)$, which makes it inefficient when $P$ is large. CG improves scalability, but slows down the convergence. The complexities of the proposed method and SM-cns are both of $\bb{\mathcal{O}}(KP)$, while their iterations are equally effective. However, as it can be seen, the proposed method is substantially faster. This is achieved by using the method explained in Section~\ref{sec: unconstrained CSC} instead of the Sherman-Morrison formula, in both the $\bb{z}$-update step (CSC phase) and the $\bb{g}$-update step (dictionary update phase).

\begin{figure}[tbhp]
\centering
{\includegraphics[width=0.5\linewidth]{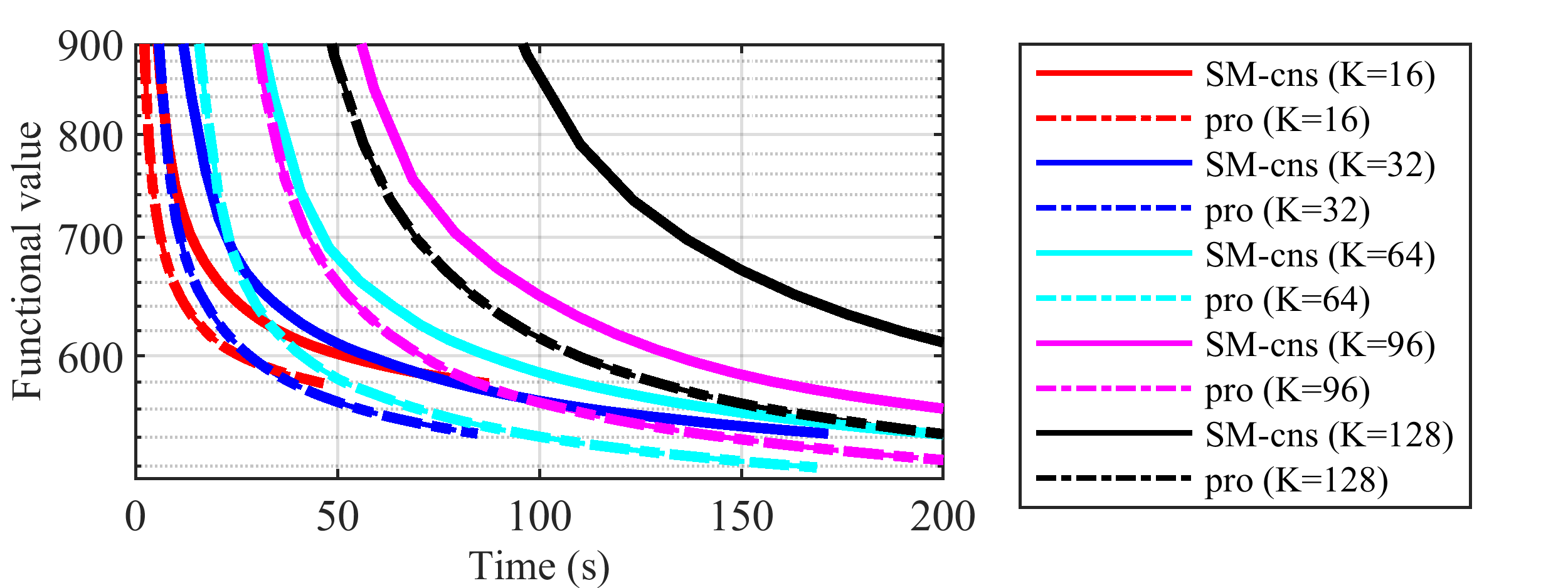}}\\
\caption{Functional values over time using different $K$ values, $P=10$, $\rho=10$, $\lambda=0.05$ and filters of size $8\times8$.}
\label{fig: CDL K comparison}
\end{figure}

In Fig.~\ref{fig: CDL K comparison} the convergence speeds of the proposed CDL method and SM-cns using different dictionary sizes ($K$) are compared. The improved computational efficiency of the proposed method can be clearly observed.
\section{Conclusion}
An efficient solution for the convolutional least-squares fitting problem has been presented. The proposed method has been used to substantially improve the efficiency of the state-of-the-art convolutional sparse coding and dictionary learning algorithms. In addition, a novel method for convolutional sparse approximation with a constraint on the approximation error has been proposed. 
\balance
\bibliographystyle{IEEEtran}
\bibliography{references}

\end{document}